\pdfoutput=1
\documentclass[11pt]{article}

\usepackage[final]{acl}

\usepackage{times}
\usepackage{latexsym}

\usepackage[T1]{fontenc}
\usepackage[utf8]{inputenc}

\usepackage{microtype}

\usepackage{inconsolata}

\usepackage{graphicx}

\usepackage{todonotes}
\usepackage[whole]{bxcjkjatype}
\usepackage{booktabs}
\usepackage{multicol}
\usepackage{multirow}
\usepackage{tcolorbox}
\usepackage{graphicx}
\author{
Shintaro Ozaki \hspace{20pt}
Kazuki Hayashi \hspace{20pt}
Miyu Oba \\ [3pt]
\textbf{Yusuke Sakai}  \hspace{15pt}
\textbf{Hidetaka Kamigaito}  \hspace{15pt}
\textbf{Taro Watanabe} \\ [4pt]
Nara Institute of Science and Technology (NAIST), Japan \\ [3pt]
  \texttt{ozaki.shintaro.ou6@naist.ac.jp} \\
  \texttt{\{sakai.yusuke.sr9, kamigaito.h, taro\}@is.naist.jp} \\
  }
\title{BQA: Body Language Question Answering Dataset \\for Video Large Language Models}

\begin{document}
\maketitle
\begin{abstract}
A large part of human communication relies on nonverbal cues such as facial expressions, eye contact, and body language. 
Unlike language or sign language, such nonverbal communication lacks formal rules, requiring complex reasoning based on commonsense understanding.
Enabling current Video Large Language Models (VideoLLMs) to accurately interpret body language is a crucial challenge, as human unconscious actions can easily cause the model to misinterpret their intent.
To address this, we propose a dataset, BQA, a body language question answering dataset, to validate whether the model can correctly interpret emotions from short clips of body language comprising 26 emotion labels of videos of body language.
We evaluated various VideoLLMs on the BQA with and without Multimodal Chain of Thought (CoT) and revealed that understanding body language is challenging, and our analyses of the wrong answers by VideoLLMs show that certain VideoLLMs made largely biased answers depending on the age group and ethnicity of the individuals. 
We also found consistent error patterns in VideoLLMs\footnote{
The dataset is available at \url{https://huggingface.co/datasets/naist-nlp/BQA}.
}.
\end{abstract}

\section{Introduction}
Video large language models (VideoLLMs)~\cite{wang2024qwen2, ye2024mplug, zhang2024llavanext-video, team2024gemini} process videos by integrating multimodal inputs into an understanding of the content.
These models take video frames, sound, and accompanying text as input and generate text, answers to questions~\cite{maaz-etal-2024-video, lei-etal-2018-tvqa}, or predictions based on the video~\cite{Xiao_2021_CVPR, yi2019clevrer}, enabling various applications, such as video summarization and question answering. 
This capability fosters a future where humans and models coexist, making it essential for VideoLLMs to grasp human emotions and body language for interaction.
One study~\cite{hyun-etal-2024-smile} has investigated emotion detection from body language, identifying smiles and their underlying causes.
However, since this approach is limited to analyzing a single emotion, it remains unclear whether the findings are generalized to all human emotions.
If VideoLLMs are unable to understand human emotion from body language, they may not be suitable for future applications such as dialogue systems and AI robots, where emotional awareness is crucial for enabling more effective interactions.

Our research focuses on the analysis of various emotional expressions in human body language.
We created a dataset called \emph{BQA}, a multiple-choice QA task, in which each body language video is associated with a question regarding a particular emotion comprising four choice answers, e.g., Surprise, Confidence, Anger, and Embarrassment, reformatting the Body Language Dataset created for pose estimation~\cite{Luo2020}.
The BQA consists of 7,632 short videos (5--10 seconds, 25 fps), depicting human body language with metadata (gender, age, ethnicity) and 26 emotion labels per video.
The BQA creation involves four steps using Gemini~\cite{team2024gemini}: extracting answer choices, generating questions, evaluating potential harm, and assigning difficulty labels.
Moreover, we evaluated recent VideoLLMs on BQA, with and without Chain of Thought (CoT)~\cite{zhang2024multimodalchainofthoughtreasoninglanguage} as well as conducted multiple human evaluations,
and found the task enough challenging for VideoLLMs. 
Error analysis revealed biases toward specific ages or ethnicities, and Multimodal CoT, despite improving accuracy, showed consistent error patterns.

\begin{figure*}[t]
    \centering
    \includegraphics[width=0.97\textwidth]{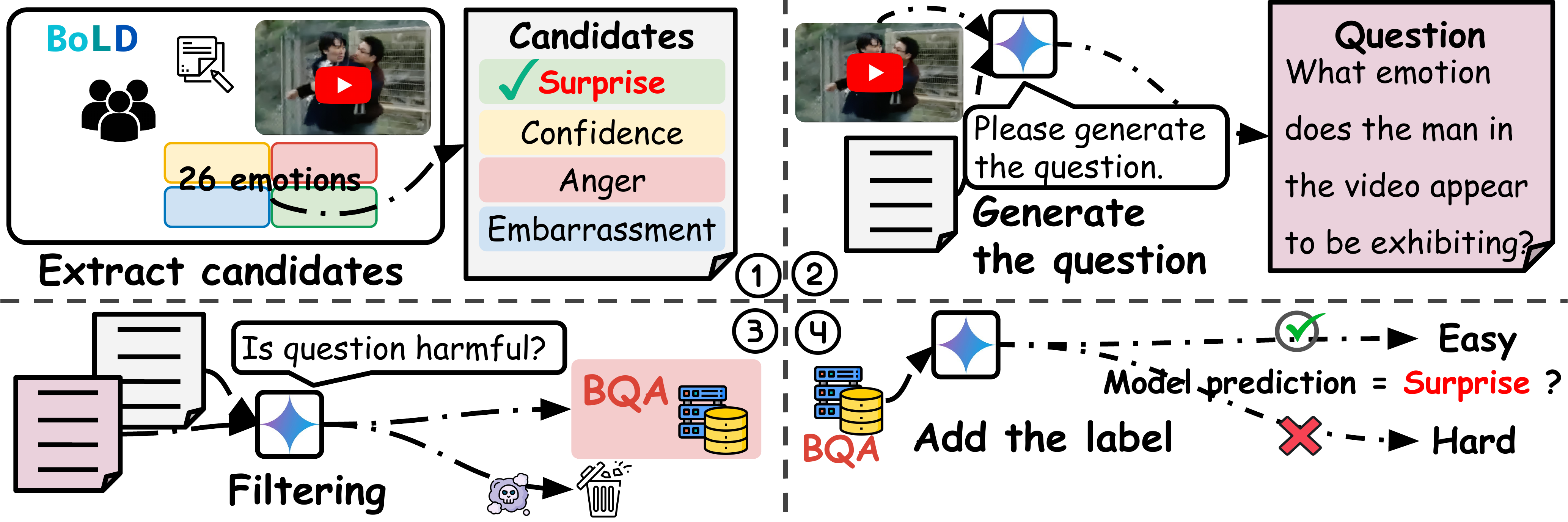}
    \caption{4 steps of creating the BQA dataset. In STEP1, candidates are created; in STEP2, questions are generated; in STEP3, filtering is conducted; and in STEP4, we assign (Easy/Hard) labels, describing the details in Section~\ref{dataset-construction}.
}
\label{fig:dataset-construction}
\end{figure*}

\section{Body Language Dataset (BoLD)}
\label{background-and-related-work}
Body Language Dataset (BoLD)~\cite{Luo2020} is a dataset for recognizing human actions and selecting appropriate emotions created by splitting 150 films, totaling 220 hours of footage, resulting in 9,876 video clips. 
Each clip is approximately 5 seconds long, comprising nearly 125 frames. These videos were annotated with 26 emotion labels~\cite{8099695} via crowdsourcing, where multiple annotators assigned emotion labels on a 10-point scale, which were normalized to represent the emotion of each video. 
Metadata, including the age, gender, and ethnicity of the individuals in the clips, is also available.
However, BoLD is designed for a model directly predicting the emotion, and is not suitable for prompting with clear answers in order to investigate the capacity of VideoLLMs, which expect inference with natural language.

\begin{figure}[t]
    \centering
    \includegraphics[width=0.98\linewidth]{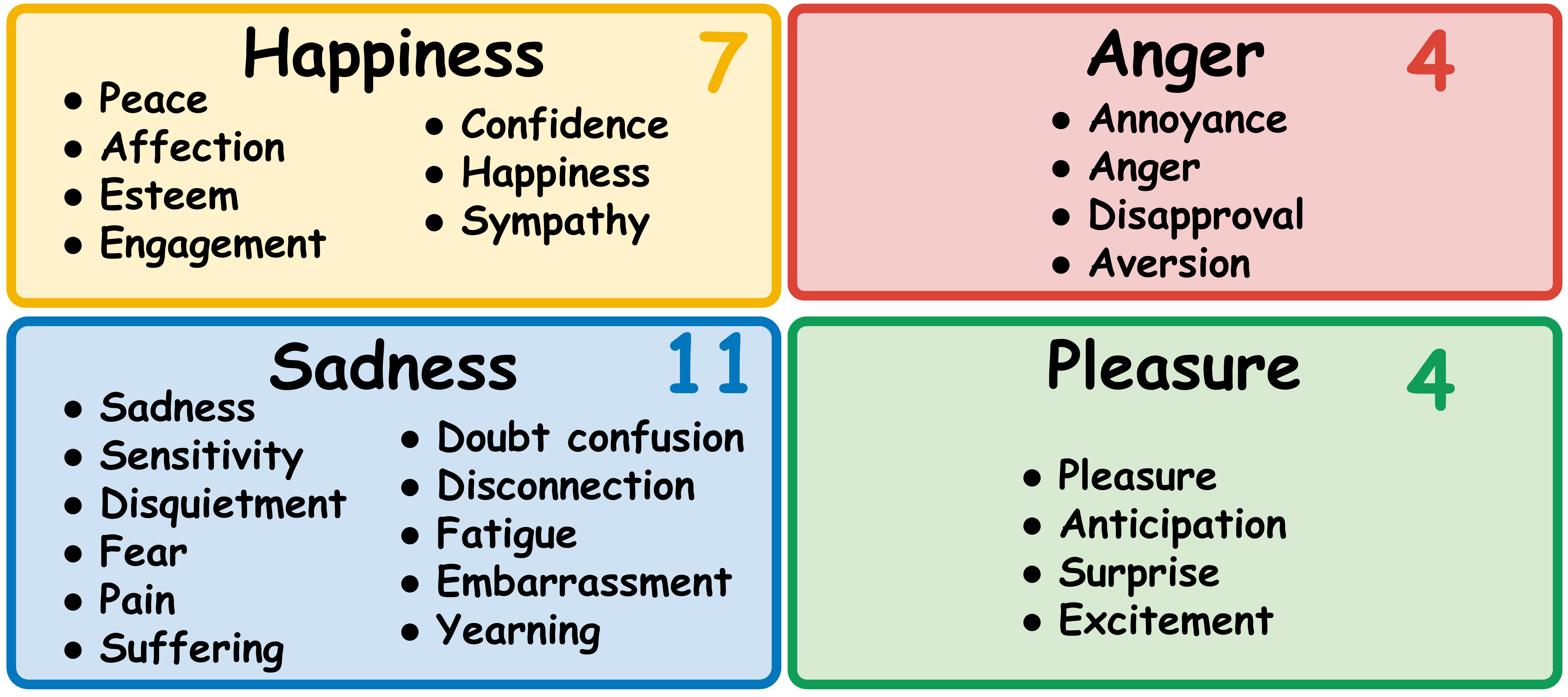}
    \caption{Categorized the 26 emotion labels into 4 groups with similar emotions to extract the candidates.}
    \label{fig:26pattern-and-4emotions}
\end{figure}

\section{Dataset Construction}
\label{dataset-construction}
We transformed BoLD into a multiple-choice QA format by generating questions from video content and using the 26 emotion labels as answers to evaluate how well VideoLLMs understand human emotions expressed through body language, adding steps to extract appropriate choices since BoLD was not designed for LLMs evaluation.
To design a QA task for evaluating the LLMs' understanding of body language for emotional expression, we followed the approach of mCSQA~\cite{sakai-etal-2024-mcsqa}, which semi-automatically generates QA questions based on candidate answers using LLMs.
The whole process consists of four steps as described in Figure~\ref{fig:dataset-construction}.
First, we extract candidate choices from the BoLD's metadata (Figure~\ref{fig:dataset-construction}-1) followed by question generation using a Gemini based on the video and the candidate choices (Figure~\ref{fig:dataset-construction}-2).
Then, we automatically filter out
 inappropriate QAs (Figure~\ref{fig:dataset-construction}-3).
Lastly, we let the Gemini solve the QAs to evaluate difficulty levels (Figure~\ref{fig:dataset-construction}-4).

\paragraph{STEP1: Extract Candidates}
We categorized 26 emotion labels defined in BoLD into 4 groups: Happiness, Anger, Sadness, and Pleasure, as shown in Figure~\ref{fig:26pattern-and-4emotions}, based on the research in which emotions are classified into four main groups~\cite{James1890-JAMPOP}.
For the creation of BQA, we apply a multiple-choice question format, where the one with the highest empathy level is treated as correct, and the remaining three options are selected from different emotion groups to ensure that the choices for the QA candidates are selected from each of the groups.
The example in Figure~\ref{fig:dataset-construction}-1 shows that the correct answer is Surprise from the Pleasure group and the remaining candidates, i.e., Confidence, Anger, and Embarrassment, are drawn from the other groups, i.e., Happiness, Anger, and Sadness, respectively.

\begin{figure}[t]
    \centering
    \includegraphics[width=0.98\linewidth]{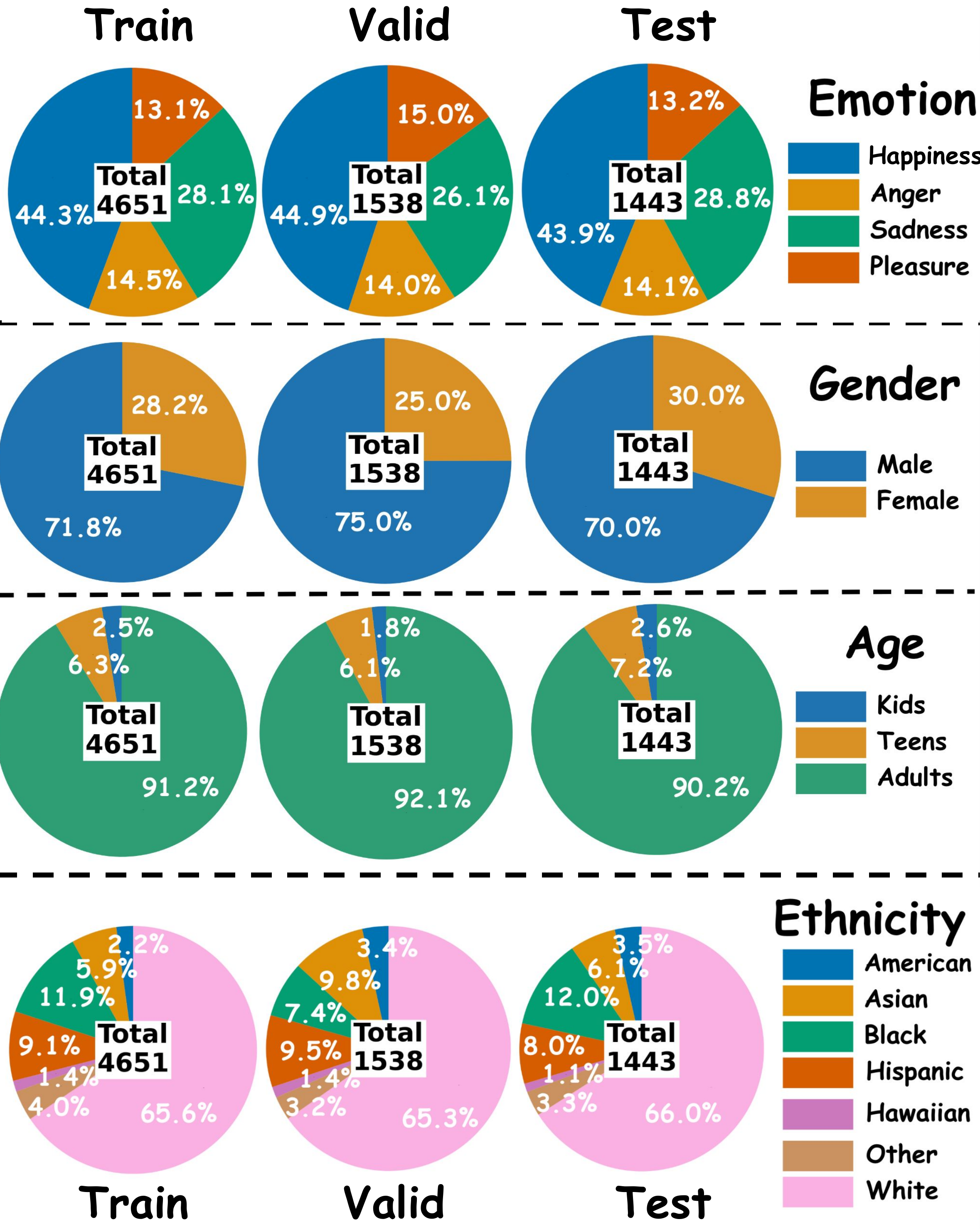}
    \caption{The proportion of metadata in the BQA.}
    \label{fig:dataset-distribution}
\end{figure}

\paragraph{STEP2: Generate the Question by VideoLLM}
Since BoLD does not follow the QA format, we create the appropriate questions from the pairs of candidates and a video by following the prompt design of mCSQA~\cite{sakai-etal-2024-mcsqa} modified for VideoLLMs.
We input the four candidate options (e.g., Confidence, Surprise, Anger, Embarrassment from each group in Figure~\ref{fig:dataset-construction}-1) along with the video into Gemini with the highest performance~\cite{team2024gemini} and let the model generate questions such as ``What emotion does the man in the video appear to be exhibiting?'' like Figure~\ref{fig:dataset-construction}-2 with the prompt we use for the generation in Appendix~\ref{prompt-on-creating-dataset}.

\paragraph{STEP3: Filter the QA by VideoLLM}
Sometimes, a VideoLLM generates a question which is relatively easy to estimate or answer, e.g., a question that already contains information about the correct candidate, such as ``The man looks so shocked. Which emotion is appropriate at this time?''
Thus, we let Gemini evaluate whether the generated questions were enough objective and whether they contained superficial information about the correct candidate.
If any outputs included the harmful content or did not conform to the conditions for the questions, they were excluded as shown in Figure~\ref{fig:dataset-construction}-3.

\paragraph{STEP4: Classify the QA as Easy or Hard}
Finally, as shown in Figure~\ref{fig:dataset-construction}-4, we let Gemini solve the created questions by labeling each question as ``Easy'' when it could answer or as ``Hard'' if it could not answer using the prompt presented in Appendix~\ref{prompt-on-creating-dataset}.
These question labels allow us to analyze whether a hard question that is not solvable by Gemini is also difficult for other VideoLLMs.
For instance, if the correct answer is ``Surprise'' and Gemini responds with ``Anger,'' then that question would be classified as ``Hard.''
After completing these four steps, we split the dataset into training, validation, and test sets in a 6:2:2 ratio, resulting in 4.5k, 1.5k, and 1.5k questions, respectively.
The precise number of data is in Table~\ref{tab:the-number-of-data}.

\paragraph{What are the Aspects that Make the Samples Easy/Hard?}
Our analysis of the issues labeled as "Hard" revealed the following findings: 

(1) Neutral expressions: Videos featuring neutral facial expressions often resulted in the hard label.

(2) Obstructions such as glasses, hats, or sunglasses: Samples involving individuals wearing such items were more likely to be misclassified, leading to the hard label.
These findings suggest that VideoLLMs heavily rely on facial expressions when interpreting body language from the footage. 

\paragraph{Dataset Analysis}
Figure~\ref{fig:dataset-distribution} shows the distribution of our datasets categorized by our emotion type and three groups of meta information in BoLD: gender, age and ethnicity annotated by humans, revealing that many of the videos feature adult males who are White. 
It displays the distribution of the four groups of emotions when the annotators selected the most appropriate emotion from the 26 patterns as the correct answer. 
While Happiness occupies a lot, the overall distribution appears to be balanced.

\section{Evaluation}
\paragraph{Experimental Setup}
The models used for evaluation include VideoLLaMA2~\cite{cheng2024videollama}, LLaVA-NeXT~\cite{zhang2024llavanext-video}, Qwen2-VL~\cite{wang2024qwen2}, and Phi-3.5~\cite{abdin2024phi} with the prompt in Appendix~\ref{gemini-configuration}.
For the test data, we also used the proprietary models, Gemini~\cite{team2024gemini} and GPT-4o~\cite{achiam2023gpt}.
We used LoRA-Tuning~\cite{hu2022lora} on VideoLLaMA2 with the configuration in Appendix~\ref{lora-configuration} using the training data and let the model answer with the correct choice in a single word.
All audio from the videos was removed to allow for evaluating the model's ability to interpret body language without relying on auditory information.
Additionally, we randomly selected 100 cases from the test set to measure human performance, describing the guideline in Appendix~\ref{instruction-for-human-eval}.
As an additional evaluation on the test set, we further employed Multimodal Chain of Thought (CoT) to make the models generate reasoning as evidence we call ``rationale'' for selecting answers.
The prompts used for these evaluations are provided in Appendix~\ref{prompt-on-creating-dataset}.

\begin{table}[t]
\centering
\small
\setlength{\tabcolsep}{2pt}
\begin{tabular}{@{}cccccccc@{}}
\toprule 
\multirow{2}{*}{VideoLLM}&\multirow{2}{*}{\#F}   & \multicolumn{3}{c}{Test} & \multicolumn{3}{c}{Test (CoT)} \\
\cmidrule(lr){3-5}\cmidrule(lr){6-8}
& & Easy & Hard & Total &  Easy & Hard & Total \\
\midrule
Human (Rand100) & - & 0.96 & 0.77 & 0.85 & - & - & - \\
\midrule
Gemini & * & 0.91 & 0.08 & \textbf{0.61} & 0.94 & 0.90 & 0.92 \\
GPT-4o & * & 0.78 & 0.38 & 0.60 & 0.97 & 0.95 & 0.96 \\
Phi-3.5 & 16 & 0.77 & 0.41 & 0.58 & 1.00 & 0.96 & \textbf{0.98} \\
Qwen2-VL & 16 & 0.68 & 0.27 & 0.47 & 0.98 & 0.95 & 0.97 \\
LLaVA-NeXT & 16 & 0.66 & 0.30 & 0.47 & 1.00 & 0.96 & \textbf{0.98} \\
VideoLLaMA2 & 16 & 0.15 & 0.01 & 0.08 & 0.15 & 0.01 & 0.08 \\
VideoLLaMA2 (FT) & 16 & 0.98 & 0.91 & 0.94 & 0.98 & 0.90 & 0.94 \\
\bottomrule
\end{tabular}

\caption{The results using BQA. \#F indicates the frame. An asterisk (*) signifies 1 fps. (FT) indicates the LoRA-Tuning model. ``Human'' is an average of 3 annotators.}
\label{tab:model-result-valid-and-test}
\end{table}

\paragraph{Main Results}
We show the results in Table~\ref{tab:model-result-valid-and-test} with the test set and Table~\ref{tab:the_result_of_valid} with the valid set.
GPT-4o and Gemini achieved higher accuracy than the other models. 
VideoLLaMA2 replied without confirming the choice of format before fine-tuning (FT), resulting in a low score, but after FT, its score surpassed Gemini's.
From this result, the label assignment in STEP4 did not cause hallucinations.
Regarding Gemini, which generated the questions, we found that the problems were sufficiently challenging even for the model itself. 
Furthermore, in STEP4 of Section~\ref{dataset-construction}, those labeled as Easy became unsolvable during inference, likely due to the prompt that restricted the output to single words.
Regarding CoT, the results showed a large improvement, confirming its effectiveness for VideoLLMs.

\section{Analysis and Discussion}
\begin{figure}[t]
    \centering    \includegraphics[width=\linewidth]{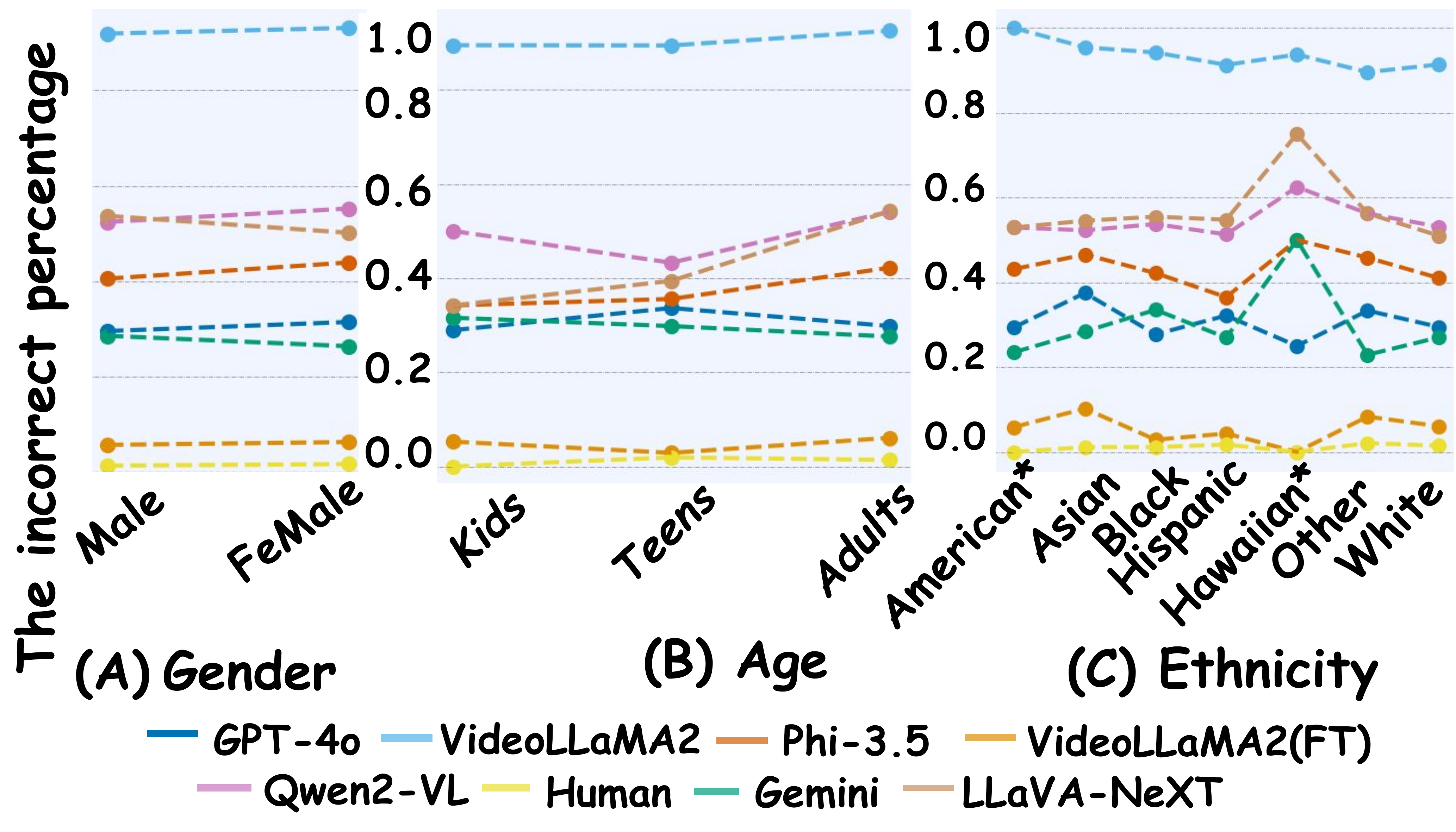}
    \caption{The analysis of incorrectly answered questions shows, from left to right, (A) gender, (B) age, and (C) ethnicity. Note that the higher value indicates more mistakes. An asterisk (*) in (C), especially ``American'' and ``Hawaiian'', indicates that they are native humankind.}
    \label{fig:output-distribution}
\end{figure}

We analyzed the videos by gender (Figure~\ref{fig:output-distribution}-A), age (Figure~\ref{fig:output-distribution}-B), and ethnicity (Figure~\ref{fig:output-distribution}-C) in which each model tends to make mistakes.
Other models showed lower evaluation results in the Hard setting compared to the Easy setting.
This indicates that even if a language model can create questions, it does not guarantee that it can solve them itself. 
Since the accuracy of the other VideoLLMs, even GPT-4o, was lower than that of Gemini, the dataset proved to be sufficiently challenging for all models.

\paragraph{Which Age do VideoLLMs Often Mistake?}
We show which age groups models tend to struggle with in Figure~\ref{fig:output-distribution}-B. Higher values indicate a greater tendency to make errors on videos featuring individuals from that age group.
These results show that most models do not exhibit bias based on age, while LLaVA-NeXT tends to make more errors on videos featuring ``Adults'' compared to the others.

\paragraph{Which Ethnicity do VideoLLMs Often Mistake?}
In Figure~\ref{fig:output-distribution}-C, we show the consistent patterns to make the mistakes based on the ethnicity of the individuals in the videos. 
Gemini and LLaVA-NeXT tend to make more errors on the problems related to ``Native Hawaiian.''
Notably, LLaVA-NeXT only achieves around 25\% accuracy on these questions.

\paragraph{Why does CoT Improve the Performance?}
While the Multimodal Chain of Thought (CoT) improved performance, we observed that the rationales often included the correct answer.
The inclusion of leaked answers contributes to the performance improvements largely.
Thus, the scores achieved by Multimodal CoT should be treated separately from the inherent difficulty of the dataset.
However, upon analyzing the types of errors even after CoT, we observed a recurring trend: many of the misclassified instances featured neutral facial expressions or minimal body movements.
This observation further supports the claim in Section~\ref{dataset-construction} that VideoLLMs tend to focus more on the face when attempting to understand the body language.

\section{Conclusion}
Our work created a dataset called BQA to evaluate whether VideoLLMs understand body language that represents emotions and let VideoLLMs solve the questions using BQA.
The results show the questions are challenging for all models, confirming the meaning of the dataset.
We analyze the types of questions each model tends to get wrong, revealing that some models show a tendency to make more mistakes based on ethnicity and age group.
The models also tend to focus more on the face than on body language, and we found that accuracy decreases when there are obstructions that obscure the face.
It is also essential to conduct evaluations that focus on the biases in VideoLLMs.

\section{Limitations}
\subsection{The Evaluation by Human Annotator}
The random sampling of 100 evaluations was conducted by three people.
The overall agreement score is 0.79, indicating a high level of agreement.
In the future, we may take care of using crowdsourcing to gather evaluations from a lot of people.
Furthermore, it would be beneficial to include evaluations from people of various ethnicities to explore differing perspectives on body language that expresses emotions.
However, \citet{tedeschi-etal-2023-whats} argues that human baselines may be unreliable due to factors such as crowdsourced worker payment issues and random sample effects. 
We should therefore be cautious about the baseline for our human evaluation.

\subsection{Video Quality}
This study takes care of the possibility that video quality or size may affect accuracy.
Specifically, BoLD uses old films, and some of them have noticeably poor quality.
Since the same data is used for evaluation, the ranking of the models' accuracy will likely remain unchanged.
However, when inputting higher-quality videos, accuracy might improve across all models.

\subsection{Frame Issues}
Although VideoLLMs claim to handle videos, many actually use image models by treating videos as a sequence of images.
In this study, we standardized the number of frames each model can process to 16, but inputting the maximum number of frames may affect the results.
However, since inputting more frames increases memory usage, we also need to be mindful of resource constraints.

\subsection{Regarding Emotional Expressions}
In this study, we categorized 26 patterns of emotional expression into 4 groups based on previous research~\cite{James1890-JAMPOP}. 
While we extracted options from these four patterns, this method may not be entirely accurate.
Future research will focus on how the models behave when we expand the available options.

\subsection{The Costs of Calling API}
The models used in this paper are GPT-4o (gpt-4o-0806) from OpenAI. 
GPT-4o is accessed via API, which is subject to change and incurs costs based on the number of input tokens. 
In this study, inference costs totaled approximately \$154 and \$100 for Multimodal CoT, but this may change in the future. 
Additionally, due to cost considerations, we used Gemini-1.5-pro. 
This model is also accessed via API, which is subject to change and incurs costs based on the number of input tokens.

\section{Ethical Considerations}
\subsection{Taking Care about Culture}
The expression of emotions through body language doesn't necessarily remain consistent across all countries. 
Therefore, we might need to update the dataset to take care of  
cultural factors in future developments.

\subsection{License}
Since BoLD does not have a clear license, we believe its use for research purposes is unproblematic. 

\subsection{Large Scale Human Evaluation}
Although human evaluations of BQA aim to minimize bias, implicit biases may still remain. 
In the future, it may be necessary to employ multiple annotators for a fair assessment. 
However, as mentioned in ~\citet{tedeschi-etal-2023-whats}, careful consideration is needed when hiring annotators, as their results may not always be accurate.

\subsection{AI Assistant Tools}
We used ChatGPT\footnote{\url{https://chatgpt.com/}} and DeepL\footnote{\url{https://www.deepl.com/ja/translator}} to translate sentences to English to accelerate our research.

\subsection{Annotators in BoLD}  
In this study, we rely on data annotated in BoLD for analysis. 
However, the annotated information may not always be accurate.
For example, a White annotator may have intentionally mislabeled an Asian person as Black. 
Additionally, implicit biases from annotators could lead to adults being mistaken for children.  

Regarding emotions, there is also a possibility of bias during the annotation process. 
We ourselves found it challenging to explain the differences between the 26 patterns of emotional expression, which is why we grouped them into four categories. 
It is unlikely that annotators fully captured these distinctions, so we must approach this with caution.

% \section*{Acknowledgement}
% We sincerely appreciate the valuable feedback from the reviewers, area chair, and senior area chair.
% \bibliography{anthology, custom}
\bibliography{custom}

%\clearpage
\appendix
\section{Additional Discussions}
\label{sec:appendix}
\subsection{Which Emotion Do VideoLLMs Often Mistake?}

\begin{figure*}[!t]
    \centering
    \includegraphics[width=0.8\linewidth]{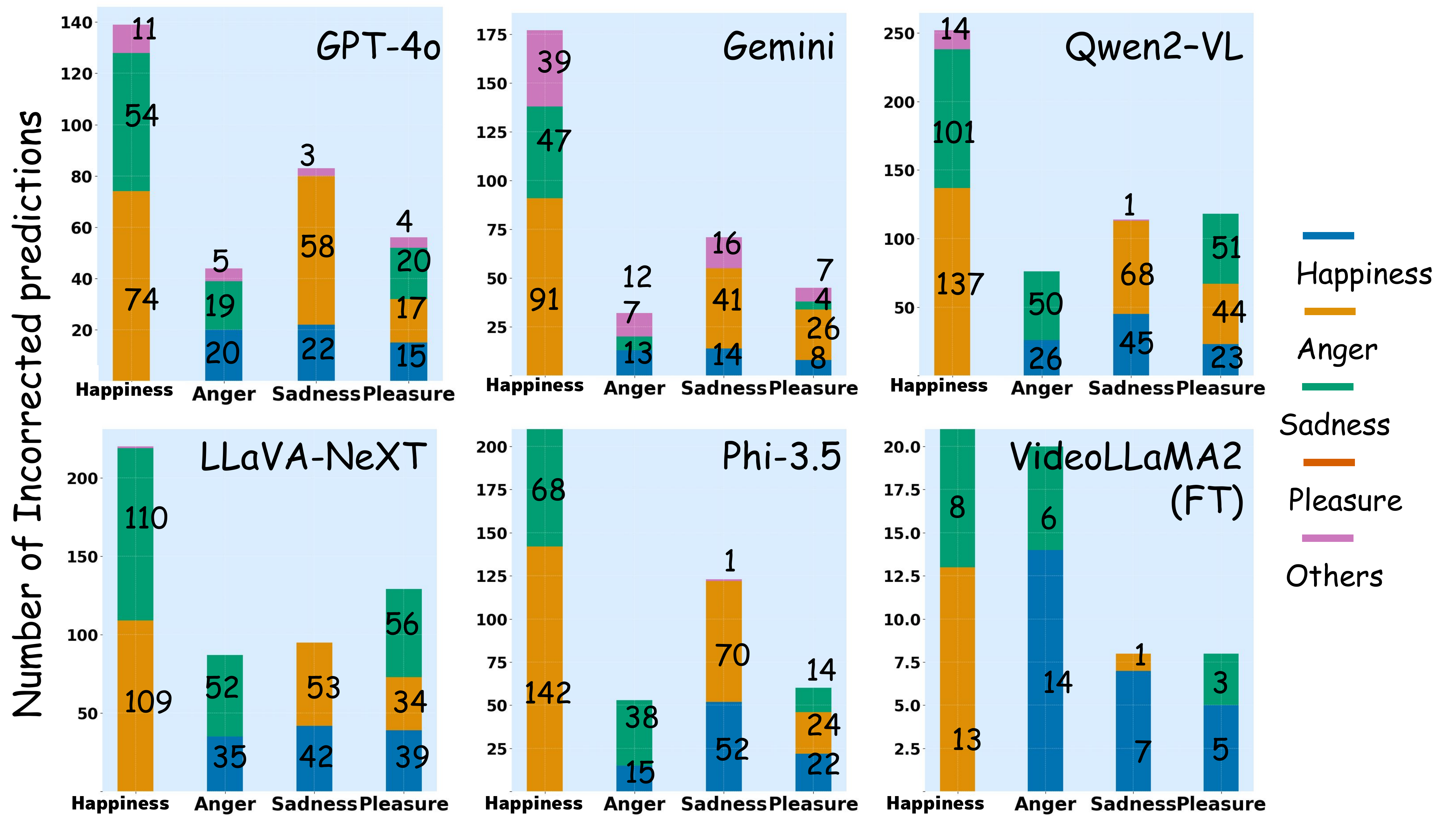}
    \caption{The emotional distribution of output from each model. The x-axis shows the label of the correct answer, and the y-axis shows how the model got it wrong in doing so.}
    \label{fig:emotional-distribution}
\end{figure*}
We analyzed which labels the models predicted for the questions they answered incorrectly in Figure~\ref{fig:emotional-distribution}. 
The x-axis represents the correct labels, while the y-axis shows the emotions predicted by the models. The ``Others'' includes instances where the model output was not an emotion, such as a sentence. 
None of the models predicted ``Pleasure'' when they made mistakes, indicating a strong capability to predict actions representing ``Pleasure.'' 
However, all models frequently failed when the correct label was ``Happiness,'' often selecting opposing emotions like ``Sadness'' or ``Anger.''

\subsection{Expansion from existing QA to Other QA Datasets}

There are several methods for extending existing QA datasets to new tasks or modalities.
For example, \citet{seo-etal-2024-kocommongen} created KoCommonGEN v2 as an expansion of Korean CommonGen~\cite{seo-etal-2022-dog}, and \citet{lynn-etal-2025-multiple} constructed a new culturally-aware dataset based on Belebele~\cite{bandarkar-etal-2024-belebele}. 
\citet{demszky2018transforming} transformed existing QA datasets, such as SQuAD~\cite{rajpurkar-etal-2016-squad} and MovieQA~\cite{tapaswi2016movieqa}, into an NLI dataset, highlighting how adapting existing resources can yield valuable contributions.
Similarly, \citet{sakajo-etal-2025-tonguescape} developed Tonguescape to benchmark VideoLLMs.
Building on this trend, we extend BoLD to construct BQA, a dataset designed to serve as input for VideoLLMs and support the understanding of body language.

\subsection{Which Gender do VideoLLMs Often Mistake?}
Figure~\ref{fig:output-distribution}-A shows the tendency of each model to make mistakes based on whether the video features a male or female subject. 
Higher values indicate a higher likelihood of errors for the videos.
From these results, we can see that none of the models exhibits bias based on gender. 
These findings suggest that the models are focused on human actions rather than whether the person is male or female.

\subsection{Trends in Generated Questions}  
As described in Section~\ref{dataset-construction}, we used Gemini, which achieved the best performance among VideoLLMs according to~\cite{fu2024video}, to generate the question texts.  
We confirmed that the generated questions predominantly begin with wh-words, i.e., \texttt{What, Where, When, Who, Why, How, Which}, which we believe is appropriate for question formation.  
Furthermore, we evaluated the diversity of the generated questions using Self-BLEU~\cite{zhu2018texygen}, and the results are shown in Table~\ref{tab:self-bleu}. These results are based on 4-gram analysis\footnote{When calculating the score, we use nltk~\cite{bird-loper-2004-nltk}.}. They indicate that the diversity of the questions is balanced across sets, confirming the fairness of the question distribution. 

\begin{table}[h!]
    \centering
    \begin{tabular}{lrr}
    \toprule
    \textbf{Set} & \textbf{Score ($\downarrow$)} & \textbf{n-gram} \\
    \midrule
    Train & 0.87 & 4 \\
    Valid & 0.86 & 4 \\
    Test  & 0.85 & 4 \\
    \bottomrule
    \end{tabular}
    \caption{The results of Self-BLEU on BQA. Lower scores indicate the more diverse questions.}
    \label{tab:self-bleu}
\end{table}

\subsection{Data Contamination}
We do not believe that BQA data has already been used for training for the following reasons.

(1): Originality of the questions: The questions used in this study were generated by the model itself during the experimental process. 
Currently, there is no scenario where the specific question-and-answer pairs created in this study would have been included in the pre-training data of proprietary models like Gemini or GPT-4o.

(2): Performance inconsistency with contamination: If data contamination had occurred, one would expect Gemini to have a significant advantage in answering the questions, as it would have been exposed to similar patterns during pre-training. 
However, as the results show, Gemini achieves only approximately 60\% accuracy on the dataset. 
This relatively modest performance strongly suggests that no data leakage has occurred. If contamination were present, the accuracy would likely be substantially higher, especially for questions generated by Gemini itself.

\subsection{Self-Preference Bias}
Self-preference bias is a phenomenon where a Large Language Model (LLM) favors its own outputs over responses from other models or human-generated texts~\cite{panickssery2024llm}.
This bias can compromise evaluation objectivity, making it essential to address its potential impact on our results.
However, in this study, only the questions were generated by the Gemini model.
The answer candidates were sourced from external datasets, i.e., BoLD, ensuring they were independent of Gemini’s generation process. 
This approach prevents any alignment-based advantage from self-generated outputs, mitigating the risk of self-preference bias.

\subsection{Exhortation of Multimodal CoT}
\label{the-scalability-of-cot}
We analyzed the large accuracy improvements achieved by applying Multimodal Chain of Thought~\cite{zhang2024multimodalchainofthoughtreasoninglanguage}. 
Upon examining the generated rationales, we observed that many explanations explicitly included reasons for selecting the correct answer while also providing reasons for not choosing the other options. 
Multimodal CoT presents the question, the answer choices, and the correct answer, then generates an explanatory rationale that derives the answer from the given choices, this rationale being subsequently used for the QA task.
These rationales often clearly incorporated the correct answer within the explanation itself, as we described below the boxes.
Unlike the traditional step-by-step reasoning approach of Chain of Thought~\cite{wei2022chain}, this explicit inclusion of the correct answer suggests a form of answer leakage, which naturally contributes to higher accuracy.  
Therefore, it is important to note that this answer leakage influences the results obtained from Multimodal CoT and may not fully reflect the inherent capabilities of the model.

\begin{tcolorbox}[title=Examples of Successful Generation about Rationale, colback=white]
\small
The man sitting in the armchair makes eye contact with the other man and sustains it while he speaks. This direct eye contact and attention to the conversation suggest active participation and engagement with what the other man is saying. 
\end{tcolorbox}

\begin{tcolorbox}[title=Examples of Bad Generation about Rationale (Data Leakage), colback=white]
\small
The speaker is looking for an address that she thought she had placed in her purse. When she realizes it is not there, she laughs and appears flustered, indicating confusion and a lack of certainty. This best aligns with the "doubt\_confusion" emotion category. The other choices are not applicable because she is not displaying affection, aversion, or disconnection.
\end{tcolorbox}

\subsection{Dataset Size}
The size of the BQA dataset after applying the four steps described in Section \ref{dataset-construction} is in Table~\ref{tab:the-number-of-data}. 
Ensuring that the dataset is sufficient to evaluate whether VideoLLMs can comprehend body language.

\begin{table}[h!]
\centering
\begin{tabular}{@{}cccc@{}}
\toprule
\textbf{Split} & \textbf{Total Size} & \textbf{Easy} & \textbf{Hard} \\ 
\midrule
Train & 4,651 & 2,192 & 2,459 \\ 
Valid & 1,538 & 746 & 792 \\ 
Test & 1,443 & 707 & 736 \\ 
\bottomrule
\end{tabular}
\caption{The data size of each split. After completing the 4 steps of dataset creation, we split the data. Our work also conducted LoRA-Tuning.}
\label{tab:the-number-of-data}
\end{table}

\subsection{The Result of Valid Set in BQA}
Table~\ref{tab:the_result_of_valid} below presents the experimental results obtained using the validation set. 
We evaluated open models using the validation set as part of the evaluation.

\begin{table}[h!]
    \small
    \centering
    \begin{tabular}{@{}ccccc@{}}
    \toprule 
    \multirow{2}{*}{VideoLLM}&\multirow{2}{*}{\#F}   & \multicolumn{3}{c}{Valid} \\
    \cmidrule(lr){3-5}
    & & Easy & Hard & Total \\
    \midrule
    Phi-3.5 & 16 & 0.76 & 0.38 & \textbf{0.56} \\
    Qwen2-VL & 16 & 0.69 & 0.32 & 0.50 \\
    LLaVA-NeXT & 16 & 0.65 & 0.31 & 0.47 \\
    VideoLLaMA2 & 16 & 0.40 & 0.09 & 0.24  \\
    VideoLLaMA2 (FT) & 16 & 0.89 & 0.68 & 0.78 \\
    \bottomrule
    \end{tabular}
    \caption{The result of valid set, \#F indicating the frame.}
    \label{tab:the_result_of_valid}
\end{table}

\section{Details of Experimental Settings}
Below, we described the details of the models evaluated in this study.
\begin{center}
\resizebox{\columnwidth}{!}{
    \begin{tabular}{@{}lll@{}}
        \toprule
        Model  & Params & HuggingFace Name  \\
        \midrule
        Qwen2-VL & 8.29B & Qwen/Qwen2-VL-7B-Instruct \\
        LLaVA-NeXT & 8.03B & lmms-lab/LLaVA-NeXT-Video-7B-Qwen2 \\
        Phi-3.5 & 4.15B & microsoft/Phi-3.5-vision-instruct \\
        VideoLLaMA2 & 8.03B & DAMO-NLP-SG/VideoLLaMA2-7B-Base \\
        GPT-4o & - & gpt-4o-2024-0806 \\
        Gemini & - & gemini-1.5-pro \\
        \bottomrule
    \end{tabular}
}
\end{center}

\subsection{LoRA Tuning Setting}
\label{lora-configuration}
We conducted LoRA~\cite{hu2022lora} tuning with the VideoLLaMA2 model. The model was trained using four NVIDIA A100-SXM4-40GB GPUs. Detailed parameters are provided in Table~\ref{tab:videollama2-lora}.

\begin{table}[h]
\centering
\footnotesize
\begin{tabular}{lc}
\toprule
Hyper Parameter & Value \\
\midrule
torch\_dtype & bfloat16 \\
seed & 42 \\
max length & 2048 \\
batch size & 4 \\
epoch & 1 \\
lora r & 128 \\
lora alpha & 256 \\
lora dropout & 0.05 \\
\multirow{2}{*}{lora target modules} & o\_proj, gate\_proj, up\_proj, v\_proj, \\ 
& q\_proj, down\_proj, k\_proj \\
\bottomrule
\end{tabular}

\caption{The hyperparameters of VideoLLaMA2~\cite{cheng2024videollama} for LoRA-Tuning~\cite{hu2022lora} used in the experiment, and others, were set to default settings~\cite{sakai-etal-2024-toward}. The implementation used the Transformers library~\cite{wolf-etal-2020-transformers}.
}
\label{tab:videollama2-lora}
\end{table}

\subsection{Gemini Settings}
\textbf{Why we use Gemini for question generation} stems from its state-of-the-art performance at the time of this study. According to ~\citet{fu2024video}, as of later in 2024, Gemini demonstrated the highest performance among multi-modal LLMs in video QA tasks. 
Consequently, Gemini was selected for question generation, as it represented the most advanced model available during the research period.

\label{gemini-configuration}
Table~\ref{tab:gemini_configuration} describes the configuration used to let the Gemini inference and generate in this study.
\begin{table}[h]
    \centering
    \resizebox{\linewidth}{!}{
    \begin{tabular}{@{}cc@{}}
        \toprule
        \textbf{Category} & \textbf{Value} \\
        \midrule
        HARM\_CATEGORY\_DANGEROUS & \multirow{5}{*}{BLOCK\_NONE} \\
        HARM\_CATEGORY\_HARASSMENT &  \\
        HARM\_CATEGORY\_HATE\_SPEECH &  \\
        HARM\_CATEGORY\_SEXUALLY\_EXPLICIT &  \\
        HARM\_CATEGORY\_DANGEROUS\_CONTENT & \\
        \bottomrule
    \end{tabular}
    }
    \caption{The configuration settings of Gemini.}
    \label{tab:gemini_configuration}
\end{table}

\subsection{Filtering by Rule-Based Algorithm}
In this study, we first performed a rule-based filtering process.
We checked if the questions ended with a question mark, whether they were a single line, and ensured that the candidates for the options were not included in the questions.
However, no instances were excluded during this filtering. 

\subsection{The Proportion of Data Split}
\begin{figure}[h]
    \centering
    \includegraphics[width=\linewidth]{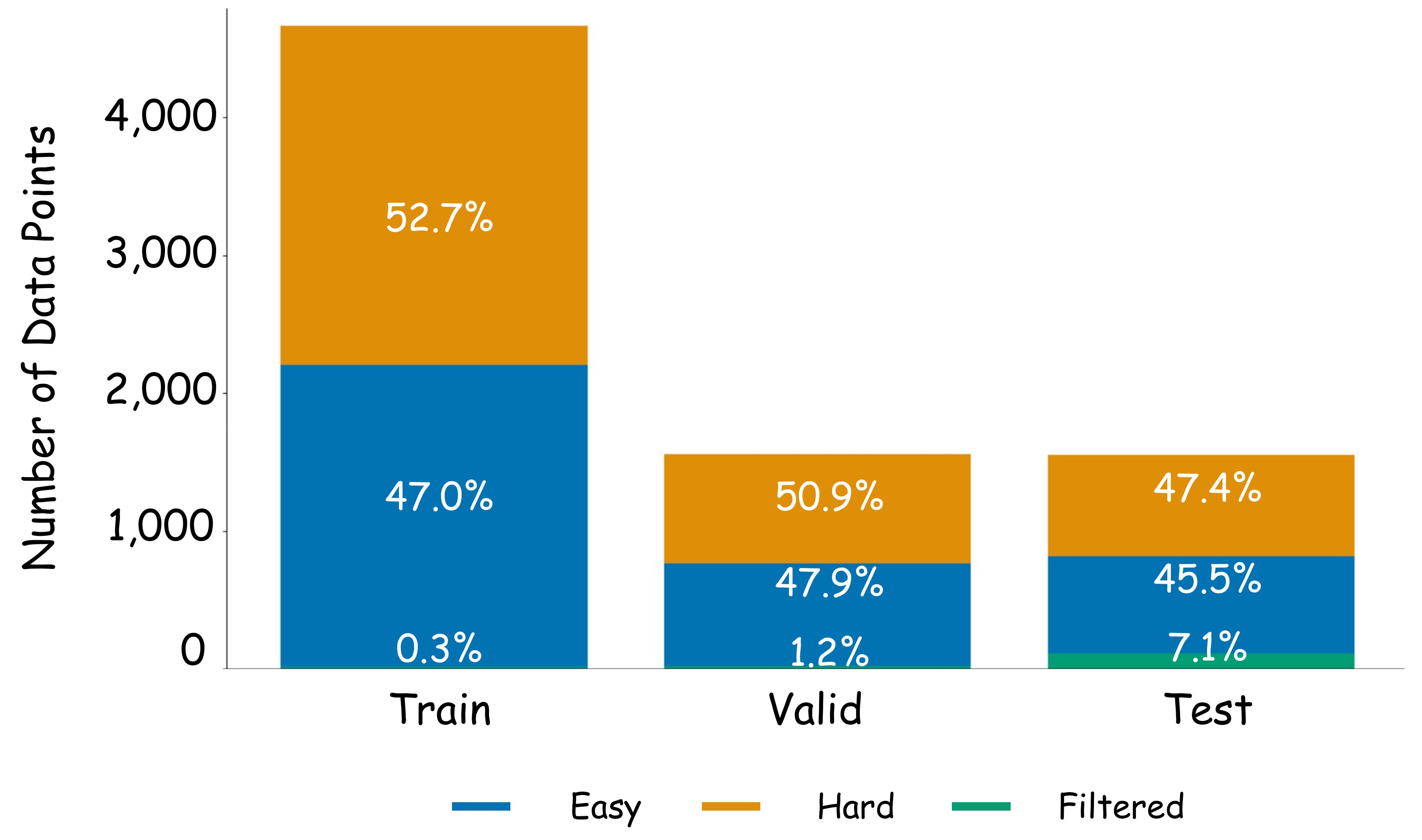}
    \caption{The percentage of each data. As stated in Section~\ref{dataset-construction}, we filtered the problem statements in STEP3 and set the difficulty levels of the problems in STEP4.}
    \label{fig:dataset-propotion}
\end{figure}

During the creation phase of BQA, if Gemini determined that a question that Gemini created was harmful, we removed it from the dataset. We then had Gemini attempt to solve the created questions, labeling them as ``Easy'' and ``Hard''. Figure~\ref{fig:dataset-propotion} shows the distribution of those labels.

\onecolumn

\section{Instruction and the Prompt}
\subsection{Instruction for Human Evaluation}
\label{instruction-for-human-eval}
We conducted human evaluations of the BQA test data. 
The instruction for requesting humans is shown below.

\begin{tcolorbox}[title=The Instruction for Human Evaluation, boxrule=1pt, colback=white]
\small
\texttt{Your task is to watch videos related to body language, read the questions, and select the appropriate option.}  \\
\texttt{Please be maximally careful about bias and try to evaluate appropriately.} \\
\texttt{Please answer with only the word of the chosen option. There are 100 questions, so take breaks as needed, and focus intently during the evaluation.} \\

\texttt{Question:} \\
\texttt{\{question\}} \\

\texttt{Choice:} \\
\texttt{\{choice0\}} \\
\texttt{\{choice1\}} \\
\texttt{\{choice2\}} \\
\texttt{\{choice3\}} \\
\end{tcolorbox}

\subsection{The Content of the Dataset}
The structure of the metadata contained in the dataset is outlined below.
The complete dataset is released at \url{https://huggingface.co/datasets/naist-nlp/BQA}.

\begin{tcolorbox}[title=An Example of Dataset, boxrule=1pt, colback=orange!20, colframe=orange!50, coltext=black, coltitle=black, colback=white]
\{ \\
　　\texttt{"video\_url": "URL\_TO\_VIDEO", \\}
　　\texttt{"answer": "affection",  \\}
　　\texttt{"question": "What feeling does the man in the video seem to express when he is smiling?", \\}
　　\texttt{"prompt": "Please look at the video entered and choose the option that applies to the following question statement.}
  \\ \\ 
　　　　\texttt{Question: \\}
　　　　\texttt{What feeling does the man in the video seem to express when he is smiling? \\ \\ }
　　　　\texttt{Choice: \\}
　　　　\texttt{affection \\}
　　　　\texttt{sympathy \\ }
　　　　\texttt{aversion \\}
　　　　\texttt{doubt\_confusion \\ \\}

　　　　\texttt{Please output only the words that apply to your answer. \\}
　　　　\texttt{If you output sentences or symbols, the answer will be incorrect.", \\}
　　\texttt{"easy\_hard\_label": "hard" \\}
\}
\end{tcolorbox}

\newpage
\subsection{The Prompt on Creating a Dataset}
\label{prompt-on-creating-dataset}
Below, we present the prompts used to instruct the model while creating the BQA. 
One prompt was for generating questions from the video and the candidates, and the other was for filtering the generated questions to determine whether they adhered to the specified conditions.

\begin{tcolorbox}[title=STEP2: The Prompt for Generate the Question, boxrule=1pt, colback=white]
\small
\texttt{First, please understand the video.} \\
\texttt{Understanding the video, please consider the candidates entered and the correct answer, and create a question for which the correct answer is the answer.} \\

\texttt{candidates:} \\
\texttt{\{candidates[0]\}} \\
\texttt{\{candidates[1]\}} \\
\texttt{\{candidates[2]\}} \\
\texttt{\{candidates[3]\}} \\

\texttt{answer: \{answer\_emotion\}} \\

\texttt{Notes:} \\
\texttt{1. Do not include the word choice in the question text.} \\
\texttt{2. Do not include superficial information such as word counts or paraphrases that can be answered by imagination.} \\
\texttt{3. Always end a sentence with a '?'.} \\
\texttt{4. Attempt to create an objective question.} \\
\texttt{5. Consist of only one sentence.}
\end{tcolorbox}

\begin{tcolorbox}[title=STEP3: The Prompt for Filtering, boxrule=1pt, colback=white]
\small
\texttt{Determine whether the input question meets the following conditions.} \\
\texttt{If it is judged to meet all five conditions, return the input sentence as it is. If it does not meet the conditions, output only the word "HARMFUL" in all uppercase letters.} \\

\texttt{The output should be either the input sentence or "HARMFUL" only.} \\

\texttt{Five conditions:} \\

\texttt{1. Does not use words from multiple-choice options.} \\
\texttt{2. Avoids using superficial information like character count, etc.} \\
\texttt{3. Ends with a question mark (?).} \\
\texttt{4. Is an objective question.} \\
\texttt{5. Consists of only one sentence.} \\
\texttt{Input sentence: \{question\}}
\end{tcolorbox}

\begin{tcolorbox}[title=STEP4: The prompt for adding the Label as Easy/Hard, boxrule=1pt, colback=white]
\small
\texttt{Question:} \\
\texttt{\{question\}} \\

\texttt{Choice:} \\
\texttt{\{choice0\}} \\
\texttt{\{choice1\}} \\
\texttt{\{choice2\}} \\
\texttt{\{choice3\}} \\

\end{tcolorbox}

\begin{tcolorbox}[title=The Prompt for Inference, boxrule=1pt, colback=white]
\small
\texttt{Please look at the video entered and choose the option that applies to the following question statement.} \\

\texttt{Question:} \\
\texttt{\{question\}} \\

\texttt{Choice:} \\
\texttt{\{choice0\}} \\
\texttt{\{choice1\}} \\
\texttt{\{choice2\}} \\
\texttt{\{choice3\}} \\

\texttt{Please output only the words that apply to your answer.} \\
\texttt{If you output sentences or symbols, the answer will be incorrect.} 
\end{tcolorbox}

\begin{tcolorbox}[title=The Prompt for generating rationale for Chain of Thought, boxrule=1pt, colback=white]
\texttt{Please look at the video entered and choose the option that applies to the following question statement.} \\

\texttt{Question:} \\
\texttt{\{question\}} \\

\texttt{Choice:} \\
\texttt{\{choice0\}} \\
\texttt{\{choice1\}} \\
\texttt{\{choice2\}} \\
\texttt{\{choice3\}} \\

\texttt{Answer: \{answer\}} \\

\texttt{Please explain the rationale to choose the correct answer.} \\
\texttt{Solution:}
\end{tcolorbox}

\begin{tcolorbox}[title=The Prompt for Inference with Chain of Thought, boxrule=1pt, colback=white]
\texttt{Please look at the video entered and choose the option that applies to the following question statement.} \\

\texttt{Question:} \\
\texttt{\{question\}} \\

\texttt{Choice:} \\
\texttt{\{choice0\}} \\
\texttt{\{choice1\}} \\
\texttt{\{choice2\}} \\
\texttt{\{choice3\}} \\

\texttt{Rationale:} \\
\texttt{\{cot\}} \\

\texttt{Answer:"""}
\end{tcolorbox}

\newpage
\subsection{Examples of BQA}
We provide several examples of actual questions and the corresponding answers from each model.
The images in the Video column capture the most distinctive moments from the videos (i.e., thumbnails). 
The complete dataset is available at \url{https://huggingface.co/datasets/naist-nlp/BQA}.
The prompt includes only the necessary parts; for the complete details, we describe in Appendix~\ref{prompt-on-creating-dataset}.

\begin{figure*}[!h]
    \centering
    \includegraphics[width=0.98\textwidth]{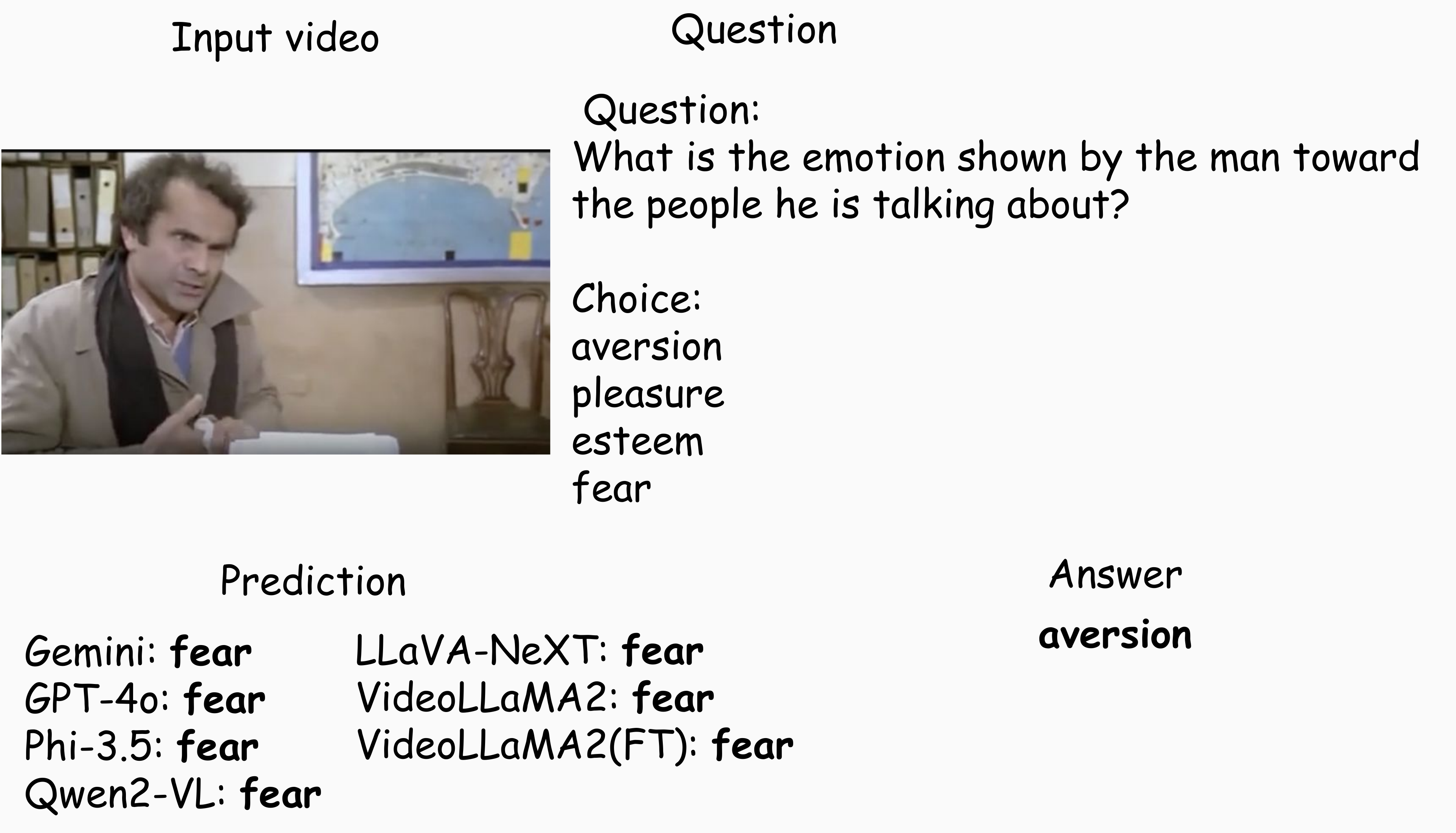}
    \caption{An example of BQA.}
    \label{fig:example1}
\end{figure*}

\begin{figure*}[!h]
    \centering
    \includegraphics[width=0.98\textwidth]{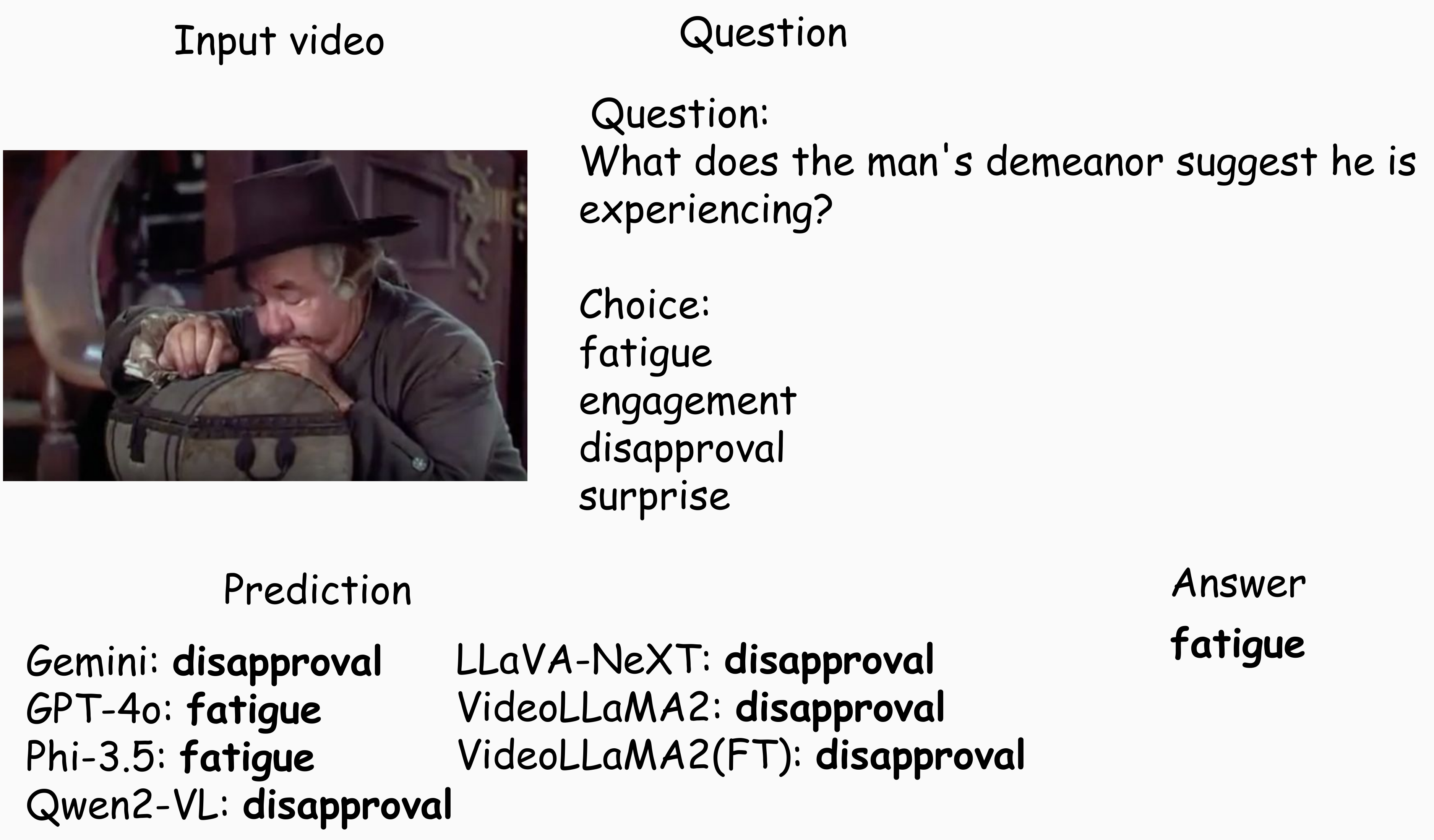}
    \caption{Another example of BQA.}
    \label{fig:example2}
\end{figure*}

\end{document}